\documentclass[10pt,twocolumn,letterpaper]{article}

\usepackage{iccv}
\usepackage{times}
\usepackage{epsfig}
\usepackage{graphicx}
\usepackage{amsmath}
\usepackage{amssymb}
\usepackage{multirow}
\usepackage{booktabs}

\usepackage[breaklinks=true,bookmarks=false]{hyperref}

\iccvfinalcopy 

\ificcvfinal\fi

\begin{document}

\title{SpiralNet++: A Fast and Highly Efficient Mesh Convolution Operator}

\author{Shunwang Gong\textsuperscript{1} \quad \quad Lei Chen\textsuperscript{2} \quad \quad Michael Bronstein\textsuperscript{1,4} \quad \quad  Stefanos Zafeiriou\textsuperscript{1,3}\\
\textsuperscript{1}Imperial College London, UK \quad \textsuperscript{2}New York University, USA \quad \textsuperscript{3}FaceSoft.io \quad \textsuperscript{4}Twitter\\
{\tt\small \{shunwang.gong16, m.bronstein, s.zafeiriou\}@imperial.ac.uk \quad lc3909@nyu.edu}
}

\maketitle
\ificcvfinal\thispagestyle{empty}\fi

\begin{abstract}
Intrinsic graph convolution operators with differentiable kernel functions play a crucial role in analyzing 3D shape meshes. In this paper, we present a fast and efficient intrinsic mesh convolution operator that does not rely on the intricate design of kernel function. We explicitly formulate the order of aggregating neighboring vertices, instead of learning weights between nodes, and then a fully connected layer follows to fuse local geometric structure information with vertex features. We provide extensive evidence showing that models based on this convolution operator are easier to train, and can efficiently learn invariant shape features. Specifically, we evaluate our method on three different types of tasks of dense shape correspondence, 3D facial expression classification, and 3D shape reconstruction, and show that it significantly outperforms state-of-the-art approaches while being significantly faster, without relying on shape descriptors. Our source code is available on GitHub\footnote{\url{https://github.com/sw-gong/spiralnet\_plus}}. 
\end{abstract}

\section{Introduction}

Geometric deep learning \cite{bronstein2017geometric} has led to a series of breakthroughs in a broad spectrum of problems ranging from biochemistry~\cite{gainza2019deciphering, gilmer2017neural}, physics~\cite{choma2018graph} to recommender systems~\cite{monti2017recommender}. This method allows computational models that are composed of multiple layers to learn representations of irregular data structures, such as graphs and meshes. The majority of current works focus on the study of generic graphs~\cite{kipf2016semi, velivckovic2018deep, xu2018powerful}, whereas it is still challenging to extract non-linear low-dimensional features from manifolds. 

A path to `solving' issues related to 3D computer vision then appears to be paved by defining intrinsic convolution operators. Attempts along this path started from formulating local intrinsic patches on meshes ~\cite{kokkinos2012intrinsic, masci2015geodesic, monti2017geometric}, and some other efforts~\cite{fey2018splinecnn, verma2018feastnet} exploit the similar idea of learning the filter weights between the nodes in a local graph neighborhood with utilizing pre-defined local pseudo-coordinate systems over the graphs.  

Driven by the significance of the design of kernel weight function, a few questions arise: \textit{Is designing better weight function the vital part of learning representations of manifolds? Can we find more efficient convolution operators without introducing elusive kernel functions and pseudo-coordinates?} It is somewhat intricate to answer if considering the problems defined on generic graphs with varied topologies. These problems, however, are possible to be addressed in terms of meshes, where data~\cite{anguelov2005scape, bogo2014faust, booth20163d, cheng20184dfab, ranjan2018generating, vlasic2008articulated} are generally aligned. 

\begin{figure}[t]
\begin{center}
   \includegraphics[width=1.0\linewidth]{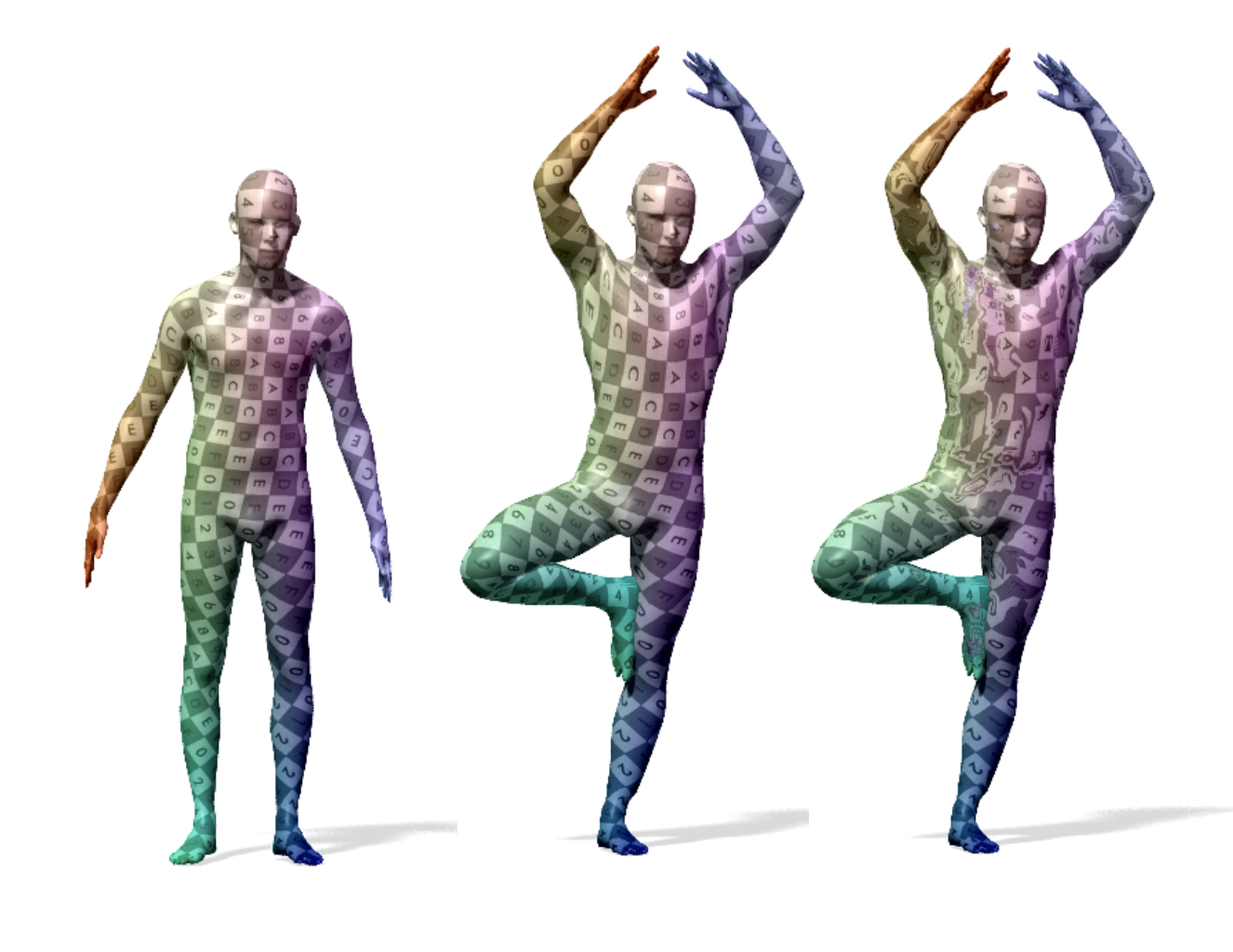}
\end{center}
\caption{Examples of texture transfer from a reference shape in neural pose (\textbf{left}) using shape correspondences predicted by SpiralNet++ (\textbf{middle}) and SpiralNet (\textbf{right})~\cite{lim2018simple}. Note that we use only 3D coordinates as input features for both methods.}
\label{fig:texture_transfer}
\end{figure}

In this paper, we address these problems by introducing a simple operator, called SpiralNet++, which captures local geometric structure from serializing the local neighborhood of vertices. Instead of randomly generating sequences per epoch \cite{lim2018simple}, SpiralNet++ generates spiral sequences \textit{only once} in order to employ the prior knowledge of fixed meshes, which improves robustness. Since our approach explicitly encodes local information, the model is capable of efficiently learning discriminative features on 3D shapes. We further propose a dilated SpiralNet++ which allows to leverage neighborhoods at multiple scales to achieve detailed captures. 

SpiralNet++ is fast, efficient, and easy to apply to various tasks in the domain of 3D computer vision. In our experiments, we bring this operator into three types of challenging problems, \textit{i.e.,} dense shape correspondence, 3D facial expression classification, and 3D shape reconstruction. Without relying on pre-processed shape descriptors or pseudo-coordinate systems, our approach outperforms the competitive baselines by a large margin in all the tasks.

\section{Related Work}

{\bf Geometric deep learning.} Geometric deep learning \cite{bronstein2017geometric} began with attempts to generalize convolutional neural networks for data with an underlying structure that is non-Euclidean. It has been widely adopted to the tasks of graphs and 3D geometry, such as node classification~\cite{kipf2016semi, velivckovic2018deep}, community detection~\cite{chen2017supervised}, molecule prediction~\cite{veselkov2019hyperfoods}, mesh deformation prediction~\cite{kostrikov2018surface}, protein interaction prediction~\cite{gainza2019deciphering}.

{\bf Dense Shape Correspondence.} We refer to related surveys~\cite{biasotti2016recent, van2011survey} on shape correspondence.
Ovsjanikov \etal~\cite{ovsjanikov2012functional} formulated a function correspondence problem to find a compact representation that could be used for point-to-point maps. 
Litany \etal~\cite{litany2017deep} took dense descriptor fields defined on two shapes as inputs and established a soft map between the two given objects, allowing end-to-end training. 
Masci \etal~\cite{masci2015geodesic} proposed to apply filters to local patches represented in geodesic polar coordinates. Boscaini \etal~\cite{boscaini2016anisotropic} proposed the ACNN by using an anisotropic patch extraction method, exploiting the maximum curvature directions to orient patches. Monti \etal~\cite{monti2017geometric} established a unified framework generalizing CNN architectures to non-euclidean domains. 
Verma \etal~\cite{verma2018feastnet} proposed a graph-convolution operator of dynamic correspondence between filter weights and neighboring nodes with arbitrary connectivity, which is computed from features learned by the network. 
Lim \etal~\cite{lim2018simple} firstly proposed SpiralNet and applied it on this task, which achieved highly competitive results. However, we observe that because spiral sequences are randomly generated at each epoch, the model is hard to converge and normally requires a larger sequence length as well as high dimensional shape descriptors as input. In order to solve these issues, we present SpiralNet++ that overcomes all of these drawbacks.

{\bf 3D Facial Expression Classification.} Facial expression recognition is a long-established computer vision problem with numerous datasets and methods having been proposed to address it. Cheng \etal~\cite{cheng20184dfab} proposed a high-resolution 4D facial expression dataset, 4DFAB, building a statistical learning model for static and dynamic expression recognition. In this paper, we are the first to introduce SpiralNet++ and other geometric deep learning methods into this task. 

{\bf Shape Reconstruction.}
Shape reconstruction is a task that recreates the surface or creates another cross-section~\cite{boissonnat1988shape}.
Ranjan \etal~\cite{ranjan2018generating} proposed a convolutional mesh autoencoder (CoMA) based on ChebyNet \cite{defferrard2016convolutional} and spatial pooling to generate 3D facial meshes. 
Bouritsas \etal~\cite{bouritsas2019neural} then integrated the idea of spiral convolution \cite{lim2018simple} into mesh autoencoder based on the architecture of CoMA, called Neural3DMM. In contrast to SpiralNet~\cite{lim2018simple}, they manually selected a reference vertex on the template mesh and defined the spiral sequence based on the shortest geodesic distance from the reference vertex. We argue that it is actually unnecessary to calculate specific spirals but only introducing redundant procedures, since under the assumption of meshes having the same topology, the spirals are already fixed and the same across all the meshes once defined. Additionally, to allow fixed-size spirals for explicit $k$-disk, they do zero-padding for the vertices that have a smaller spiral length than the average length of $k$-disk. Intuitively, vertices with a shorter spiral sequence than the average would decrease training efficiency of the weights applied on the concatenated feature vectors, since non-negligible zero paddings always have them not updated. In this paper, our approach addresses these deficiencies and shows the state-of-the-art performance on this task.

\section{Our Approach}

We assume the input domain is represented as a manifold triangle mesh $\mathcal{M}=(\mathcal{V}, \mathcal{E}, \mathcal{F})$, where $\mathcal{V},\mathcal{E},\mathcal{F}$ correspond to sets of vertices, edges and faces.

\subsection{Main Concept}
In contrast to previous approaches \cite{fey2018splinecnn, monti2017geometric, verma2018feastnet} which aggregate neighboring node features based on trainable weight functions, our method encodes node features under a explicitly defined spiral sequence, and a fully connected layer follows to encode input features combined with ordering information. It is a simple yet efficient approach. In the following sections, we will elaborate on the definition of spiral sequence and the convolution operation in detail.

\subsection{Spiral Sequence}

\begin{figure}[t]
\begin{center}
   \includegraphics[width=1.0\linewidth]{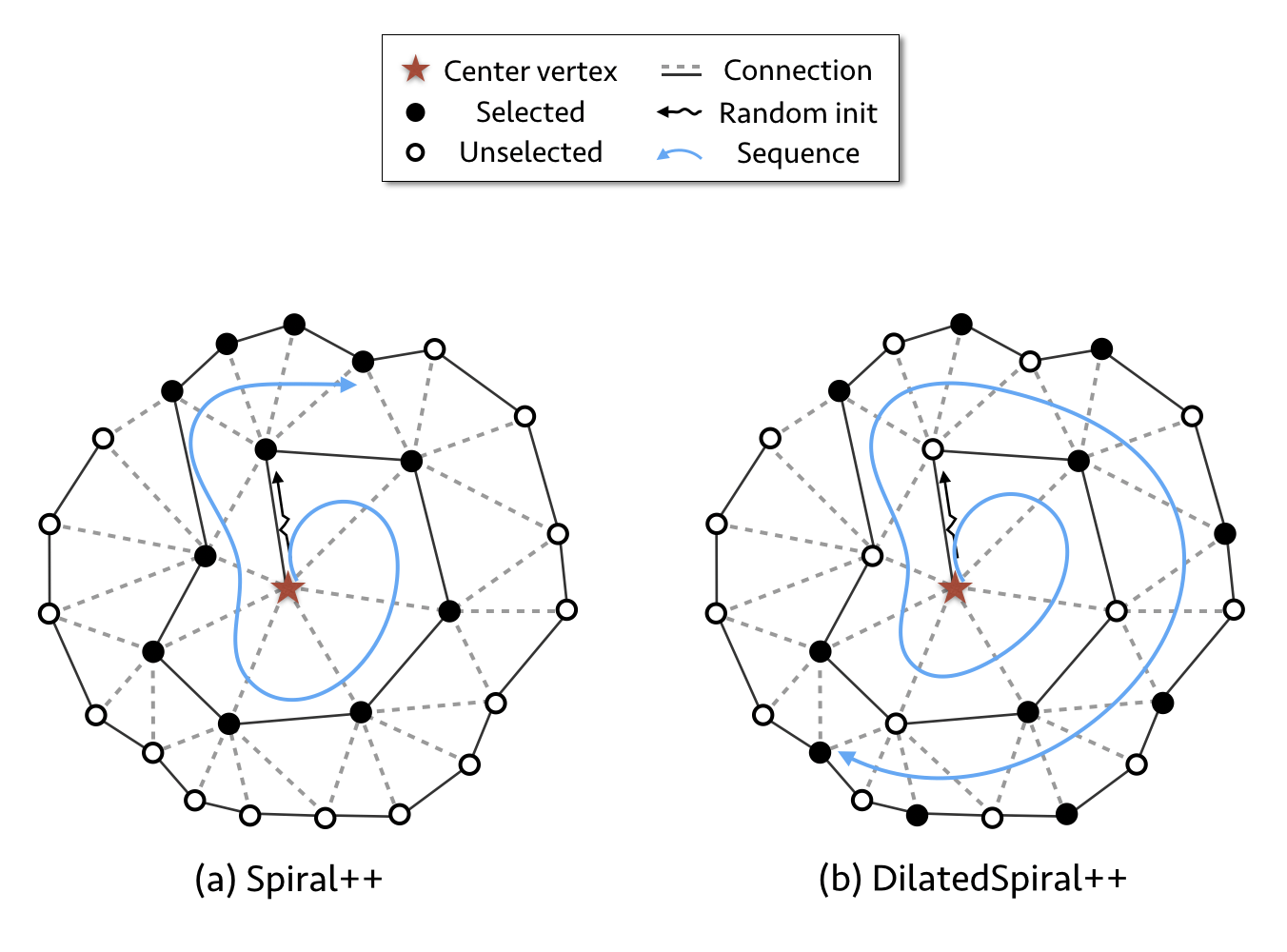}
\end{center}
\caption{Examples of Spiral++ and DilatedSpiral++ on a triangle mesh. Note that the dilated version supports exponential expansion of the receptive field without increasing the spiral length.}
\label{fig:spiral++}
\end{figure}

We begin with the definition of spiral sequences, which is the core step of our proposed operator. Given a center vertex, the sequence can be quite naturally enumerated by intuitively following a spiral, as illustrated in Figure \ref{fig:spiral++}. The degrees of freedom are merely the orientation within each ring (clockwise or counter-clockwise) and the choice of the starting direction. We fix the orientation to counter-clockwise here and choose an \textit{arbitrary} starting direction. The spirals are pre-computed \textit{only once}.

We first define a $k$-ring and a $k$-disk around a center vertex $v$ as follows:
\begin{align*}
    0\mathrm{-ring}(v) &= \{v\}, \\
    k\mathrm{-disk}(v) &= \cup_{i=0,\dots,k} i\mathrm{-ring}(v), \\
    (k+1)\mathrm{-ring}(v) &= \mathcal{N}(k\mathrm{-ring}(v))\backslash k\mathrm{-disk}(v), 
\end{align*}
\noindent where $\mathcal{N}(V)$ is the set of all vertices adjacent to any vertex in set $V$.

Here we denote the spiral length as $l$. Then $\mathrm{S}(v,l)$ is an \textit{ordered} set consisting of $l$ vertices from a concatenation of $k$-rings. Note that only part of the last ring will be concatenated to ensure a fixed-length serialization. We define it as follows:
\begin{align*}
    \mathrm{S}(v,l) \subset (0\mathrm{-ring}(v), 1\mathrm{-ring}(v), \dots , k\mathrm{-ring}(v)).
\end{align*}
It shows remarkable advantages to allow the model to learn a high-level feature representation in terms of each vertex in a consistent and robust way when we \textit{freeze} spirals during training. Compared with SpiralNet~\cite{lim2018simple}, we credit the major improvement of our approach in terms of speed and efficiency to employing the nature of aligned meshes. Note that since we do not restrict spirals to the scope of a predefined number of rings, we are not involved in performance decays caused by introducing zero-padding~\cite{bouritsas2019neural}. Furthermore, under the assumption of meshes having the same topology, the same vertex across meshes will always have the same spiral sequence regardless of the choice of starting direction, which eases the pain of manually defining the reference point and calculating the start point. By serializing the local neighborhood of vertices we are able to encode relevant information in a straightforward way with very little preprocessing. 

\subsection{Spiral Convolution}

An euclidean CNN~\cite{lecun1989backpropagation} designs a two-dimensional kernel sliding on 2D images and maps $D$ input feature maps to $E$ output feature maps. 

A common extension of CNNs into irregular domains, such as graphs, is typically expressed as a \textit{neighborhood aggregation} or \textit{message passing} scheme. With $\mathbf{x}^{(k-1)}_i \in \mathbb{R}^F$ denoting node features of node $i$ and $\mathbf{e}^{(k-1)}_{i, j} \in \mathbb{R}^{D}$ denoting (optional) edge features from node $i$ to node $j$ in layer $(k-1)$, message passing graph neural networks can be described as:
\begin{align*}
    \mathbf{x}^{(k)}_i = \gamma^{(k)} \left( \mathbf{x}^{(k-1)}_i, \square_{j \in \mathcal{N}(i)} \, \phi^{(k)}(\mathbf{x}_i^{(k-1)}, \mathbf{x}_j^{(k-1)},\mathbf{e}^{(k-1)}_{i,j}) \right)
\end{align*}
\noindent where $\mathbf{x}_i^{(k)} \in \mathbb{R}^{F'}$, and $\square$ denotes a differentiable permutation-invariant function, \textit{e.g.}, sum, mean or max, and $\phi$ denotes a differentiable kernel function. $\gamma$ represents MLPs. In contrast to CNNs for regular inputs, where there is a clear one-to-one mapping, the main challenge in the case of irregular domains is to define the corerspondence between neighbors and weight matrices which relies on  the kernel function $\phi$.

Thanks to the nature of the spiral serialization of neighboring nodes, we can define our spiral convolution in an equivalent manner to the euclidean CNNs, easing the pain of calculating the assignment value of $\mathbf{x}_j$ to the weight matrix. We define our spiral convolution operator for a node $i$ as
\begin{align*}
    \mathbf{x}^{(k)}_i = \gamma^{(k)} \left( \underset{j \in \mathrm{S}(i, l)}{||} \mathbf{x}^{(k-1)}_j \right)
\end{align*}
\noindent where $\gamma$ denotes MLPs and $\mathbin\Vert$ is the concatenation operation. Note that we concatenate node features in the spiral sequence following the order defined in $\mathrm{S}(i, l)$.

\paragraph{Dilated spiral convolution.} With the motivation of exponentially expanding the receptive field without losing resolution or coverage, we define dilated spiral convolution operators. Obviously, spiral convolution operators could immediately gain the power of capturing multi-scale contexts without increasing complexity from uniformly sampling the spiral sequence while keeping the same spiral length, as illustrated in Figure \ref{fig:spiral++}. 

\begin{figure*}[ht]
\begin{center}
   \includegraphics[width=1.0\linewidth]{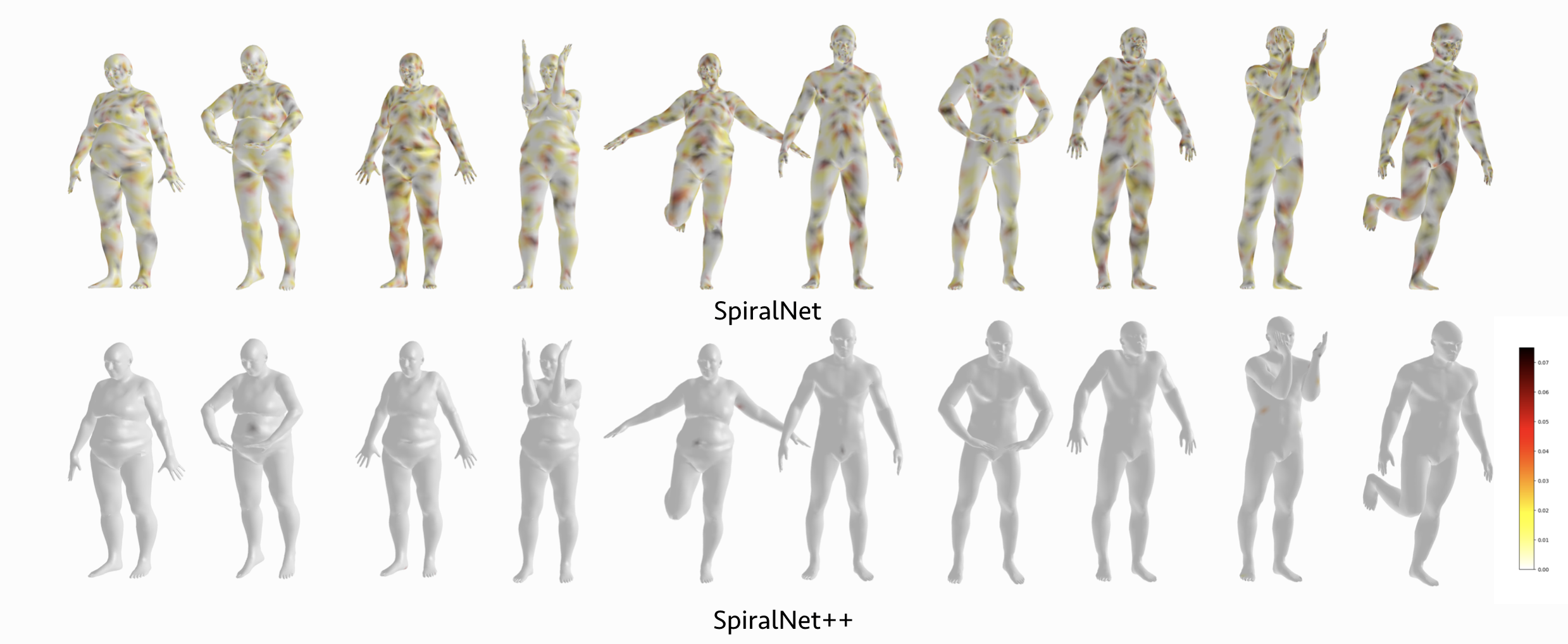}
\end{center}
\caption{Visualization of pointwise geodesic errors (in \% geodesic diameter) of our method and SpiralNet \cite{lim2018simple} on the test shapes of the FAUST~\cite{bogo2014faust} human dataset. The error values are saturated at 7.5\% of the geodesic diameter, which corresponds to approximately 15 cm. Hot colors represent large errors.}
\label{fig:vis_correspondence}
\end{figure*}

\section{Experiments}

In this section, we evaluate our method on three tasks, \textit{i.e.,} dense shape correspondence, 3D facial expression classification, and 3D shape reconstruction. We compare our method against FeaStNet \cite{verma2018feastnet}, MoNet \cite{monti2017geometric}, ChebyNet \cite{defferrard2016convolutional} and SpiralNet \cite{lim2018simple}. To enable a fair comparison, the model architectures and the kernel size of different convolutions are the same and fixed, which yields the same level of parameterization. Furthermore, we use raw 3D coordinates as input node features instead of 3D shape descriptors as traditionally used for shape analysis. All the compared methods are with our implementation in order to enforce the same experimental setting except for Neural3DMM \cite{bouritsas2019neural} that we utilize their code directly. We train and evaluate each method on a single NVIDIA RTX 2080 Ti. 

\begin{table}
\begin{center}
\begin{tabular}{c|c|c|c}
\toprule
Method & Acc. (\%) & Time/Epoch & \# Param   \\
\midrule
FeaStNet \cite{verma2018feastnet} & 79.24 & 3.016s & 1.91M   \\
MoNet \cite{monti2017geometric} & 86.05 & 1.962s & 1.91M   \\
ChebyNet \cite{defferrard2016convolutional} & 98.77 & 2.634s & 1.91M   \\
\midrule
SpiralNet \cite{lim2018simple} & 72.84 & 2.756s & 1.91M   \\
SpiralNet-LSTM \cite{lim2018simple} & 25.15 & 3.653s & 1.93M   \\
\midrule
SpiralNet++ & \textbf{99.88} & \textbf{0.98}s & 1.91M   \\
SpiralNet-LSTM++ & 97.86  & 1.989s & 1.93M  \\
\bottomrule
\end{tabular}
\end{center}
\caption{\textbf{Dense shape correspondence} on the FAUST~\cite{bogo2014faust} dataset. Test accuracy is the ratio of the correct correspondence prediction with the geodesic error of 0.}
\label{table:shape_correspondence}
\end{table}

\begin{figure}[ht]
\begin{center}
   \includegraphics[width=1.0\linewidth]{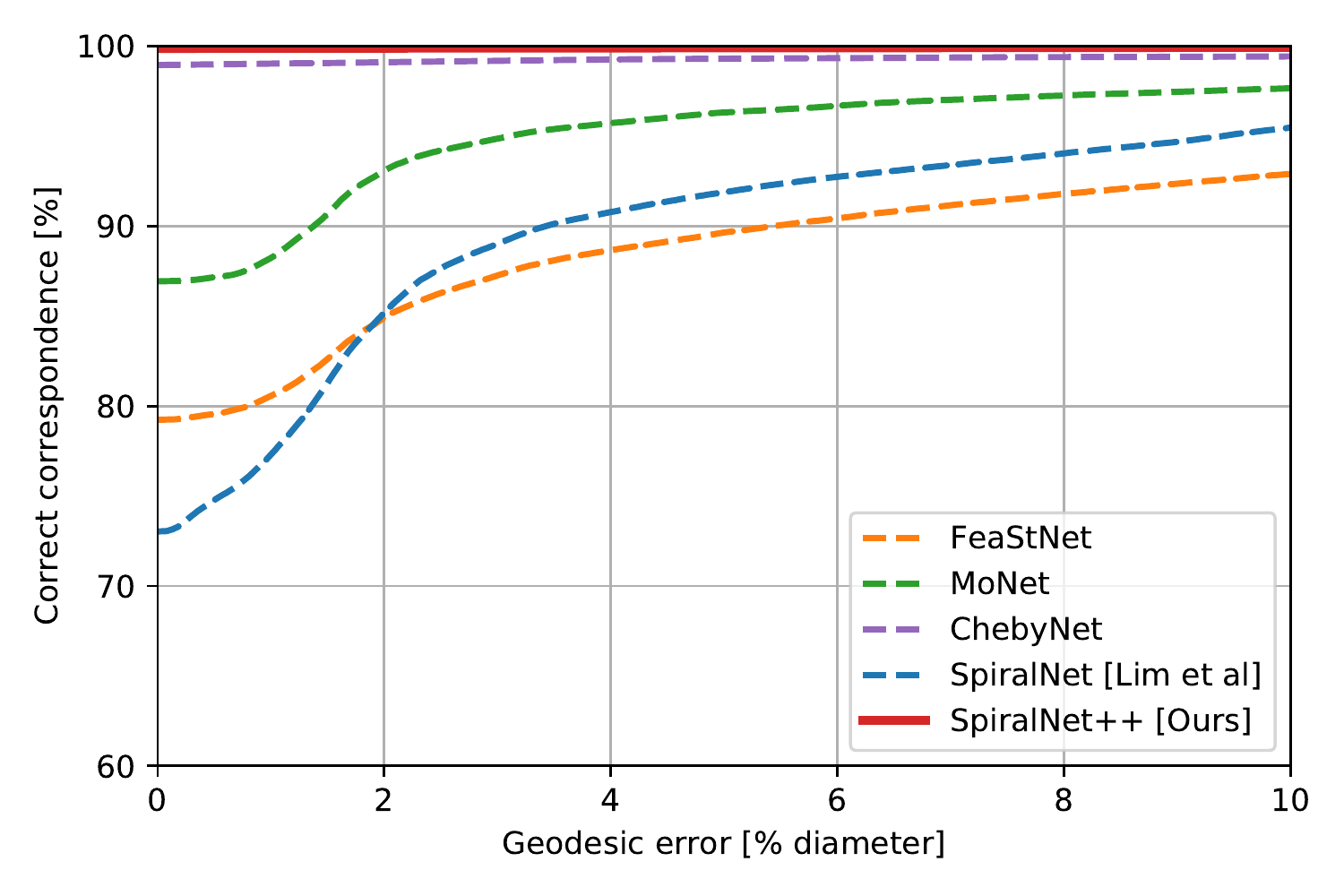}
\end{center}
   \caption{Geodesic error plot of the \textbf{shape correspondence} experiments on the FAUST \cite{bogo2014faust} humans dataset. Geodesic error is measured according to the Princeton benchmark protocol \cite{kim2011blended}. The $x$ axis displays the geodesic error in \% of diameter and the $y$ axis shows the percentage of correspondences that lie within a given geodesic error around the correct node.}
\label{fig:plot_error_shape_correspondence}
\end{figure}

\subsection{Dense Shape Correspondence}

We validate our method on a collection of three-dimensional meshes solving the task of shape correspondence similar to \cite{boscaini2016learning, masci2015geodesic, monti2017geometric, verma2018feastnet}. Shape correspondence refers to the task of labeling each node of a given shape to the corresponding node of a reference shape \cite{masci2015geodesic}. We use the FAUST dataset \cite{bogo2014faust}, containing 10 scanned human shapes in 10 different poses, resulting in a total of 100 non-watertight meshes with 6,890 nodes each. The first 80 subjects in FAUST were used for training with the remaining 20 for testing. 

\begin{table*}[ht]
\begin{center}
\begin{tabular}{c|c|c|c|c|c|c|c|c|c}
\toprule
Methods & Anger & Disgust & Fear & Happiness & Sadness & Surprise & Acc. (\%) & Time/Epoch & \# Param  \\
\midrule
Baseline \cite{cheng20184dfab}  & - & - & - & - & - &  - & 70.27 & - & - \\
\midrule
FeaStNet \cite{verma2018feastnet} & 48.40 & 70.47 & 63.43 & 85.86 & 64.00 & 92.06 & 69.89 $\pm$ 1.43 & 7.364s & 69.6k  \\
MoNet \cite{monti2017geometric} & 57.87 & 73.41 & 63.29 & 82.86 & 59.29 & 89.52  & 70.29 $\pm$ 3.55 & 6.457s & 69.5k  \\
ChebyNet \cite{defferrard2016convolutional} & 65.47 & 73.18 & 71.14 & \textbf{92.00} & 67.41 & 90.48 & 75.85 $\pm$ 1.47 & 6.009s & 69.4k  \\
\midrule
SpiralNet++  & \textbf{68.40} & \textbf{82.47} & \textbf{71.57} & 91.29 & \textbf{67.65} & \textbf{93.97} & \textbf{78.59} $\pm$ \textbf{0.64} & \textbf{3.604}s & 69.4k  \\
\bottomrule
\end{tabular}
\end{center}
\caption{\textbf{3D facial expression classification} on the 4DFAB~\cite{cheng20184dfab} facial expression dataset. We present the test accuracies obtained by all the methods for each expression (\textit{i.e.,} anger, disgust, fear, happiness, sadness and surprise) and all the expressions. *As for the Baseline~\cite{cheng20184dfab}, we use the reported result in their paper.}
\label{table: face_classification}
\end{table*}

\paragraph{Architectuers and parameters.}
As for all the experiments, we follow the network architecture of \cite{masci2015geodesic}. It consists of the following sequence of linear layers (1x1 convolutions) and graph convolutions: Lin(16)$\rightarrow$Conv(32)$\rightarrow$Conv(64)$\rightarrow$Conv(128)$\rightarrow$Lin(256)\\
$\rightarrow$Lin(6890), where the numbers indicate the amount of output channels of each layer. A non-linear activation function, ELU (exponential linear unit), is used after each Conv and the first linear layer. The kernel size or spiral length of all the Convs is 10.  

The models are trained with the standard cross-entropy classification loss. We take Adam~\cite{kingma2014adam} as the optimizer with the learning rate of 3e-3 (SpiralNet++, MoNet, ChebyNet), 1e-3 (SpiralNet), 1e-2 (FeaStNet), and dropout probability 0.5. As for input features we use the raw 3D XYZ vertice coordinates instead of 544 dimensional SHOT descriptors which was previously used in MoNet \cite{monti2017geometric}, SpiralNet \cite{lim2018simple}.

\paragraph{Discussion.} 
In Table \ref{table:shape_correspondence}, we present the accuracy of the exact correspondence (with $0\%$ geodesic error) obtained by SpiralNet++ and other approaches. It shows that our method significantly outperforms all the baselines with 99.88\% accuracy and it's counterpart SpiralNet. It should be noted that our method enjoys an extremely fast speed with the training time of 0.98s per epoch in average, which owes to our method exploiting the essence of the fixed mesh topologies. From experiments, We also observed that SpiralNet~\cite{lim2018simple} generally requires around 2500 epochs to converge while it is sufficient for SpiralNet++ to converge within 100 epochs. In Figure \ref{fig:plot_error_shape_correspondence}, we plot the percentage of correspondences that are within a certain geodesic error. In Figure \ref{fig:vis_correspondence}, it can be seen that most nodes are classified correctly with our method, which is much better than SpiralNet. Figure \ref{fig:texture_transfer} visualizes the obtained correspondence using texture transfer.

\subsection{3D Facial Expression Classification}

As the second experiment, we address the problem of 3D facial expression classification using the 4DFAB dataset \cite{cheng20184dfab}, which is a large scale dataset of high-resolution 3D faces. Previous efforts against this task focused on extracting low-dimensional features with PCA and LDA based on manually defined facial landmarks and a multi-class SVM was then employed to classify expressions \cite{cheng20184dfab}. Similar to the deep convolutional neural networks used to classify the high-resolution images in the ImageNet \cite{krizhevsky2012imagenet}, we develop an end-to-end hierarchical architecture with our method and other geometric deep learning approaches (\textit{e.g.,} ChebyConv \cite{defferrard2016convolutional}, FeaStConv \cite{verma2018feastnet}, MoNet \cite{monti2017geometric}) to solve this 3D mesh classification problem. Following the experimental setup introduced in \cite{cheng20184dfab}, we partition the data into 10 folds, and 17 distinct participants in testset are not shown in trainset (with 153 distinct participants). The number of each class is balanced in both training set and test set.

\paragraph{Pooling.} The models use a mesh pooling operation based on edge contraction \cite{garland1997surface}. The pooling operation iteratively contracts vertex pairs to simplify meshes, while maintaining surface error approximation using quadric metrics. The output feature is then directly obtained by the multiplication of input feature with a downsampling transform matrix. We denote a pooling layer using this algorithm with Pool($c$), with $c$ being the downsampling factor.

\begin{figure*}[htp]
\begin{center}
  \includegraphics[width=0.95\linewidth]{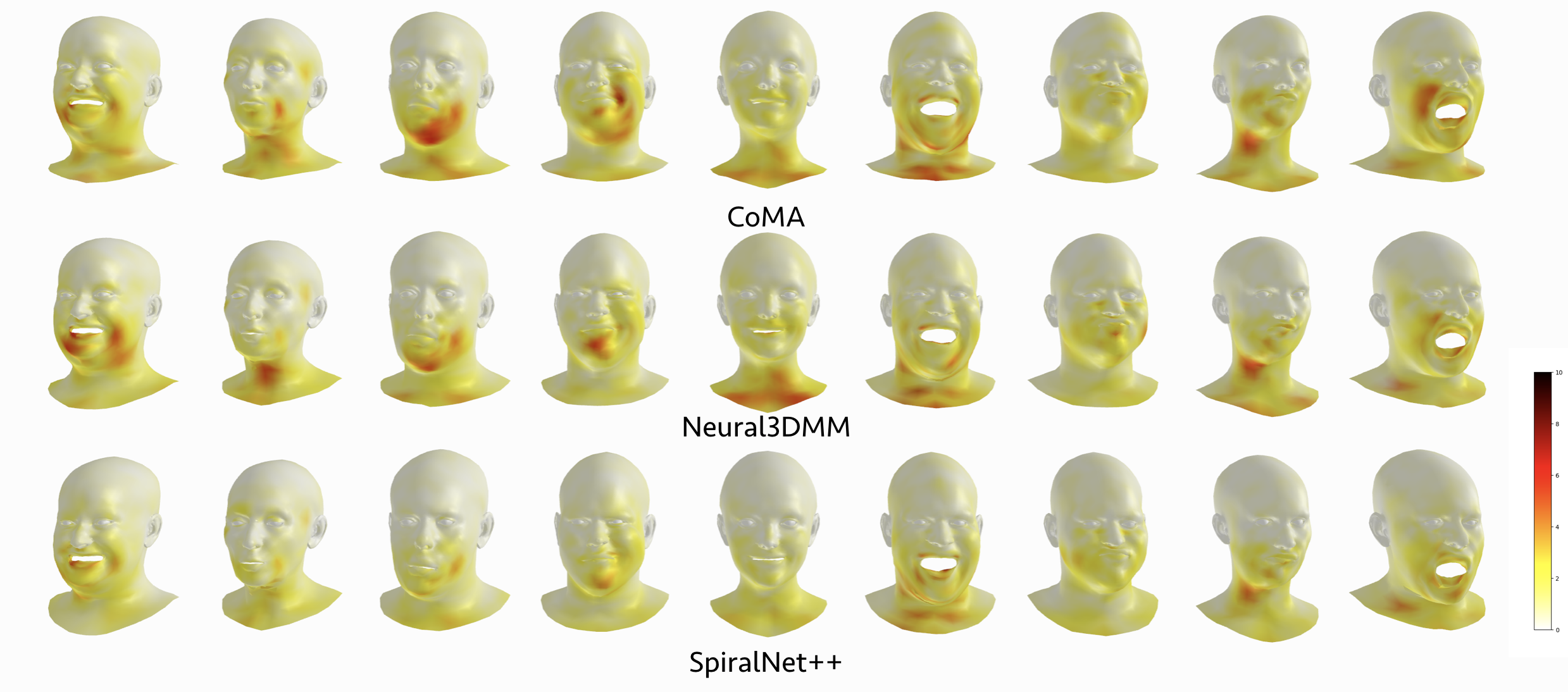}
\end{center}
  \caption{Qualitative results of 3d shape reconstruction in the CoMA \cite{ranjan2018generating} dataset. Pointwise error (euclidean distance from the groundtruth) is computed for visualization. The error values are saturated at 10 (millimeters). Hot colors represent large errors.}
\label{fig:vis_reconstruction}
\end{figure*}

\begin{table*}
\begin{center}
\begin{tabular}{c|c|c|c|c}
\toprule
Method & Mean Error & Median Error & Time/Epoch & \# Params  \\
\midrule
FeaStNet \cite{verma2018feastnet} & 0.523 $\pm$ 0.643 & 0.297 & 133.183s & 157.9k  \\
MoNet \cite{monti2017geometric} & 0.526 $\pm$ 0.605 & 0.353 & 97.009s & 155.4k  \\
\midrule
CoMA \cite{ranjan2018generating} & 0.470 $\pm$ 0.598 & 0.263 & 77.943s & 117.5k  \\
ChebyConv (K=9) \cite{defferrard2016convolutional} & 0.436 $\pm$ 0.562 & 0.242 & 86.627s & 154.9k  \\
Neural3DMM \cite{bouritsas2019neural} & 0.443 $\pm$ 0.560 & 0.245 & 107.137 & 157.0k  \\
\midrule
SpiralNet++ & \textbf{0.426 $\pm$ 0.538} & \textbf{0.238} & \textbf{30.417}s & 154.9k  \\
DilatedSpiralNet++ & \textbf{0.423 $\pm$ 0.534} & \textbf{0.236} & \textbf{29.181}s & 154.9k  \\
\bottomrule
\end{tabular}
\end{center}
\caption{\textbf{3D shape reconstruction} experiments results in the CoMA \cite{ranjan2018generating} dataset. Errors are in millimeters.}
\label{table:shape_reconstruction}
\end{table*}

\paragraph{Architectures and parameters.} \label{Pool} We design the following end-to-end architecture to classify 3D facial expressions: Conv(16) $\rightarrow$ Pool(4) $\rightarrow$ Conv(16) $\rightarrow$ Pool(4) $\rightarrow$ FC(32) $\rightarrow$ FC(6). Dropout with a probability of 0.5 is used before each FC layer. We take a standard cross entropy loss function and ELU activation function. Training is done for 300 epochs with the learning rate of 1e-3, learning rate decay of 0.99 per epoch, L2 regularization of 5e-4, batch size of 32. 

It should be noted that raw 3D XYZ coordinates are used as the input, and for MoNet, we use the relative Cartesian coordinates of linked nodes as its pseudo-coordinates. Furthermore, we fixed the same hyperparameters (\textit{i.e.,} kernel size, spiral sequence length or order of the polynomial K) for each convolution, which gives the same size of parameter space of $\mathbb{R}^{K\times \mathrm{C}_{\mathrm{in}} \times \mathrm{C}_\mathrm{out}}$ in terms of each convolution layer. 

\paragraph{Discussion}
All results of the 3D facial expression classification are shown in Table \ref{table: face_classification}. It shows that with our proposed architecture, all of the graph convolution operations outperform the baseline \cite{cheng20184dfab}. We credit these improvements to the capacity of learning intrinsic shape features compared to the baseline method. Specifically, our method achieves the highest recognition rate of 78.59\% on average. This indicates that SpiralNet++ can be successfully applied to \textit{multi-scale mesh data} improving previous results in this domain. Furthermore, it can be seen that our method is much more faster than all the other approaches.

\subsection{3D Shape Reconstruction}

As our largest experiment, we evaluate the effectiveness of SpiralNet++ on an extreme facial expression dataset. We demonstrate that a standard autoencoder architecture with SpiralNet++ allows the synthesis of high-fidelity 3D face with rich expression details. We use the dataset introduced in \cite{ranjan2018generating}, which consists of 12 classes of extreme expressions, containing over 20,465 3D meshes, each with about 5,023 vertices and 14,995 edges. Following the interpolation experimental setup \cite{bouritsas2019neural, ranjan2018generating}, we divide the dataset into training and test sets with a split ratio of 9:1. We compare our SpiralNet++ against a number of baselines including CoMA \cite{ranjan2018generating} and Neural3DMM \cite{bouritsas2019neural}, and furthermore, for the first time, we bring MoNet \cite{monti2017geometric} and FeaStNet \cite{verma2018feastnet} into this problem to explore the performance of other intrinsic convolution operations on generative models. It is worth highlighting that in the original work of CoMA \cite{ranjan2018generating}, they used ChebyNet with $K=6$. However, in order to have a fair comparison with other experiments, we show both results obtained with $K=6$ (\textit{i.e.,} CoMA) and $K=9$. In the end, we evaluate our proposed dilated spiral convolution on this problem.

\paragraph{Pooling and unpooling.} The performance of each generative model is closely related to the pooling and unpooling procedures. The same pooling strategy introduced in Section \ref{Pool} is used here. In the unpooling stage, contracted vertices are recovered using the barycentric coordinates of the closet triangle in the decimated mesh~\cite{ranjan2018generating}.

\paragraph{Architectures and parameters.} We build a standard autoencoder architecture, consisting of an encoder and a decoder. The encoder includes several convolutional layers interleaved between pooling layers, and one fully connected layer is applied in the end of the encoder to encode non-linear mesh representations. Specifically, the structure is: 3 $\times$ \{Conv(32)$\rightarrow$ Pool(4)\} $\rightarrow$ \{Conv(64) $\rightarrow$ Pool(4)\} $\rightarrow$ FC(16), with ELU activation function after each Conv layer. The structure of the decoder is the reversed order of the encoder with the replacement of pooling layers to unpooling layers. Note that one more convolutional layer with the output dimensional of 3 should be added to the end of the decoder to reconstruct 3D shape coordinates. Training is done using Adam~\cite{kingma2014adam} for 300 epochs with learning rate of 0.001, learning rate decay of 0.99 per epoch and a batch size of 32. 

We evaluate all the methods with the same architecture and hyperparameters. The kernel size of each methods is set as 9 in order to keep aligned with Neural3DMM \cite{bouritsas2019neural}, where they chose \textit{1}-hop deriving the spiral length of 9.

\paragraph{Discussion.} Table \ref{table:shape_reconstruction} shows mean euclidean errors with standard deviations, median errors and the training time per epoch. Our SpiralNet++ and its dilated version outperform all the other approaches. The result of our proposed dilated spiral convolution validates our assumption, which shows the higher capacity of capturing non-linear low-dimensional representations of 3D shape meshes without increasing parameters. We credit this improvement to its larger receptive field brought by sampling larger input feature space. Moreover, we should stress the remarkable speed of our method. With the same autoencoder architecture, SpiralNet++ is a few times faster than all the other methods. It should be noted that the performance of Neural3DMM is even worse than CoMA when bring weight matrices to the same number, which can be attributed to the fact that model learning is disrupted from introducing non-negligible information (\textit{i.e.}, zero-padding). The performance of Neural3DMM would decrease with the variance of vertex degrees increase. Figure \ref{fig:vis_reconstruction} shows the visualization of reconstructed faces in the test set. Larger errors can be seen from the faces generated by CoMA and Neural3DMM, and in particular, it become worse on the faces with extreme expressions. However, SpiralNet++ shows better reconstruction quality in these cases.

\section{Conclusions}

We explicitly introduce SpiralNet++ to the domain of 3D shape meshes, where data are generally aligned instead of varied topologies, which allows SpiralNet++ to efficiently fuse neighboring node features with local geometric structure information. We further apply this method to the tasks of dense shape correspondence, 3D facial expression classification and 3D shape reconstruction. Extensive experimental results show that our approach are faster and outperform competitive baselines in all the tasks.

\section{Acknowledgements}

SG and MB were supported in part by the ERC Consolidator Grant No.724228 (LEMAN), Google
Faculty Research Awards, Amazon AWS Machine Learning Research grant, and the Royal Society Wolfson Research Merit award. SZ was partially supported by the EPSRC Fellowship DEFORM (EP/S010203/1) and a Google Faculty Award.

{\small
\bibliographystyle{ieee}
\bibliography{egbib}

\begin{thebibliography}{10}\itemsep=-1pt

\bibitem{anguelov2005scape}
D.~Anguelov, P.~Srinivasan, D.~Koller, S.~Thrun, J.~Rodgers, and J.~Davis.
\newblock Scape: shape completion and animation of people.
\newblock In {\em ACM transactions on graphics (TOG)}, volume~24, pages
  408--416. ACM, 2005.

\bibitem{biasotti2016recent}
S.~Biasotti, A.~Cerri, A.~Bronstein, and M.~Bronstein.
\newblock Recent trends, applications, and perspectives in 3d shape similarity
  assessment.
\newblock In {\em Computer Graphics Forum}, volume~35, pages 87--119. Wiley
  Online Library, 2016.

\bibitem{bogo2014faust}
F.~Bogo, J.~Romero, M.~Loper, and M.~J. Black.
\newblock Faust: Dataset and evaluation for 3d mesh registration.
\newblock In {\em Proceedings of the IEEE Conference on Computer Vision and
  Pattern Recognition}, pages 3794--3801, 2014.

\bibitem{boissonnat1988shape}
J.-D. Boissonnat.
\newblock Shape reconstruction from planar cross sections.
\newblock {\em Computer vision, graphics, and image processing}, 44(1):1--29,
  1988.

\bibitem{booth20163d}
J.~Booth, A.~Roussos, S.~Zafeiriou, A.~Ponniah, and D.~Dunaway.
\newblock A 3d morphable model learnt from 10,000 faces.
\newblock In {\em Proceedings of the IEEE Conference on Computer Vision and
  Pattern Recognition}, pages 5543--5552, 2016.

\bibitem{boscaini2016learning}
D.~Boscaini, J.~Masci, E.~Rodol{\`a}, and M.~Bronstein.
\newblock Learning shape correspondence with anisotropic convolutional neural
  networks.
\newblock In {\em Advances in Neural Information Processing Systems}, pages
  3189--3197, 2016.

\bibitem{boscaini2016anisotropic}
D.~Boscaini, J.~Masci, E.~Rodol{\`a}, M.~M. Bronstein, and D.~Cremers.
\newblock Anisotropic diffusion descriptors.
\newblock In {\em Computer Graphics Forum}, volume~35, pages 431--441. Wiley
  Online Library, 2016.

\bibitem{bouritsas2019neural}
G.~Bouritsas, S.~Bokhnyak, M.~Bronstein, and S.~Zafeiriou.
\newblock Neural 3d morphable models: Spiral convolutional networks for 3d
  shape representation learning and generation.
\newblock {\em arXiv preprint arXiv:1905.02876}, 2019.

\bibitem{bronstein2017geometric}
M.~M. Bronstein, J.~Bruna, Y.~LeCun, A.~Szlam, and P.~Vandergheynst.
\newblock Geometric deep learning: going beyond euclidean data.
\newblock {\em IEEE Signal Processing Magazine}, 34(4):18--42, 2017.

\bibitem{chen2017supervised}
Z.~Chen, X.~Li, and J.~Bruna.
\newblock Supervised community detection with line graph neural networks.
\newblock {\em arXiv preprint arXiv:1705.08415}, 2017.

\bibitem{cheng20184dfab}
S.~Cheng, I.~Kotsia, M.~Pantic, and S.~Zafeiriou.
\newblock 4dfab: A large scale 4d database for facial expression analysis and
  biometric applications.
\newblock In {\em Proceedings of the IEEE conference on computer vision and
  pattern recognition}, pages 5117--5126, 2018.

\bibitem{choma2018graph}
N.~Choma, F.~Monti, L.~Gerhardt, T.~Palczewski, Z.~Ronaghi, P.~Prabhat,
  W.~Bhimji, M.~Bronstein, S.~Klein, and J.~Bruna.
\newblock Graph neural networks for icecube signal classification.
\newblock In {\em 2018 17th IEEE International Conference on Machine Learning
  and Applications (ICMLA)}, pages 386--391. IEEE, 2018.

\bibitem{defferrard2016convolutional}
M.~Defferrard, X.~Bresson, and P.~Vandergheynst.
\newblock Convolutional neural networks on graphs with fast localized spectral
  filtering.
\newblock In {\em Advances in neural information processing systems}, pages
  3844--3852, 2016.

\bibitem{fey2018splinecnn}
M.~Fey, J.~Eric~Lenssen, F.~Weichert, and H.~M{\"u}ller.
\newblock Splinecnn: Fast geometric deep learning with continuous b-spline
  kernels.
\newblock In {\em Proceedings of the IEEE Conference on Computer Vision and
  Pattern Recognition}, pages 869--877, 2018.

\bibitem{gainza2019deciphering}
P.~Gainza, F.~Sverrisson, F.~Monti, E.~Rodola, M.~M. Bronstein, and B.~E.
  Correia.
\newblock Deciphering interaction fingerprints from protein molecular surfaces.
\newblock {\em bioRxiv}, page 606202, 2019.

\bibitem{garland1997surface}
M.~Garland and P.~S. Heckbert.
\newblock Surface simplification using quadric error metrics.
\newblock In {\em Proceedings of the 24th annual conference on Computer
  graphics and interactive techniques}, pages 209--216. ACM
  Press/Addison-Wesley Publishing Co., 1997.

\bibitem{gilmer2017neural}
J.~Gilmer, S.~S. Schoenholz, P.~F. Riley, O.~Vinyals, and G.~E. Dahl.
\newblock Neural message passing for quantum chemistry.
\newblock In {\em Proceedings of the 34th International Conference on Machine
  Learning-Volume 70}, pages 1263--1272. JMLR. org, 2017.

\bibitem{kim2011blended}
V.~G. Kim, Y.~Lipman, and T.~Funkhouser.
\newblock Blended intrinsic maps.
\newblock In {\em ACM Transactions on Graphics (TOG)}, volume~30, page~79. ACM,
  2011.

\bibitem{kingma2014adam}
D.~P. Kingma and J.~Ba.
\newblock Adam: A method for stochastic optimization.
\newblock {\em arXiv preprint arXiv:1412.6980}, 2014.

\bibitem{kipf2016semi}
T.~N. Kipf and M.~Welling.
\newblock Semi-supervised classification with graph convolutional networks.
\newblock {\em arXiv preprint arXiv:1609.02907}, 2016.

\bibitem{kokkinos2012intrinsic}
I.~Kokkinos, M.~M. Bronstein, R.~Litman, and A.~M. Bronstein.
\newblock Intrinsic shape context descriptors for deformable shapes.
\newblock In {\em 2012 IEEE Conference on Computer Vision and Pattern
  Recognition}, pages 159--166. IEEE, 2012.

\bibitem{kostrikov2018surface}
I.~Kostrikov, Z.~Jiang, D.~Panozzo, D.~Zorin, and J.~Bruna.
\newblock Surface networks.
\newblock In {\em Proceedings of the IEEE Conference on Computer Vision and
  Pattern Recognition}, pages 2540--2548, 2018.

\bibitem{krizhevsky2012imagenet}
A.~Krizhevsky, I.~Sutskever, and G.~E. Hinton.
\newblock Imagenet classification with deep convolutional neural networks.
\newblock In {\em Advances in neural information processing systems}, pages
  1097--1105, 2012.

\bibitem{lecun1989backpropagation}
Y.~LeCun, B.~Boser, J.~S. Denker, D.~Henderson, R.~E. Howard, W.~Hubbard, and
  L.~D. Jackel.
\newblock Backpropagation applied to handwritten zip code recognition.
\newblock {\em Neural computation}, 1(4):541--551, 1989.

\bibitem{lim2018simple}
I.~Lim, A.~Dielen, M.~Campen, and L.~Kobbelt.
\newblock A simple approach to intrinsic correspondence learning on
  unstructured 3d meshes.
\newblock In {\em Proceedings of the European Conference on Computer Vision
  (ECCV)}, pages 0--0, 2018.

\bibitem{litany2017deep}
O.~Litany, T.~Remez, E.~Rodola, A.~Bronstein, and M.~Bronstein.
\newblock Deep functional maps: Structured prediction for dense shape
  correspondence.
\newblock In {\em Proceedings of the IEEE International Conference on Computer
  Vision}, pages 5659--5667, 2017.

\bibitem{masci2015geodesic}
J.~Masci, D.~Boscaini, M.~Bronstein, and P.~Vandergheynst.
\newblock Geodesic convolutional neural networks on riemannian manifolds.
\newblock In {\em Proceedings of the IEEE international conference on computer
  vision workshops}, pages 37--45, 2015.

\bibitem{monti2017geometric}
F.~Monti, D.~Boscaini, J.~Masci, E.~Rodola, J.~Svoboda, and M.~M. Bronstein.
\newblock Geometric deep learning on graphs and manifolds using mixture model
  cnns.
\newblock In {\em Proceedings of the IEEE Conference on Computer Vision and
  Pattern Recognition}, pages 5115--5124, 2017.

\bibitem{monti2017recommender}
F.~Monti, M.~Bronstein, and X.~Bresson.
\newblock Geometric matrix completion with recurrent multi-graph neural
  networks.
\newblock In {\em Advances in Neural Information Processing Systems}, pages
  3697--3707, 2017.

\bibitem{ovsjanikov2012functional}
M.~Ovsjanikov, M.~Ben-Chen, J.~Solomon, A.~Butscher, and L.~Guibas.
\newblock Functional maps: a flexible representation of maps between shapes.
\newblock {\em ACM Transactions on Graphics (TOG)}, 31(4):30, 2012.

\bibitem{ranjan2018generating}
A.~Ranjan, T.~Bolkart, S.~Sanyal, and M.~J. Black.
\newblock Generating 3d faces using convolutional mesh autoencoders.
\newblock In {\em Proceedings of the European Conference on Computer Vision
  (ECCV)}, pages 704--720, 2018.

\bibitem{van2011survey}
O.~Van~Kaick, H.~Zhang, G.~Hamarneh, and D.~Cohen-Or.
\newblock A survey on shape correspondence.
\newblock In {\em Computer Graphics Forum}, volume~30, pages 1681--1707. Wiley
  Online Library, 2011.

\bibitem{velivckovic2018deep}
P.~Veli{\v{c}}kovi{\'c}, W.~Fedus, W.~L. Hamilton, P.~Li{\`o}, Y.~Bengio, and
  R.~D. Hjelm.
\newblock Deep graph infomax.
\newblock {\em arXiv preprint arXiv:1809.10341}, 2018.

\bibitem{verma2018feastnet}
N.~Verma, E.~Boyer, and J.~Verbeek.
\newblock Feastnet: Feature-steered graph convolutions for 3d shape analysis.
\newblock In {\em Proceedings of the IEEE Conference on Computer Vision and
  Pattern Recognition}, pages 2598--2606, 2018.

\bibitem{veselkov2019hyperfoods}
K.~Veselkov, G.~Gonzalez, S.~Aljifri, D.~Galea, R.~Mirnezami, J.~Youssef,
  M.~Bronstein, and I.~Laponogov.
\newblock Hyperfoods: Machine intelligent mapping of cancer-beating molecules
  in foods.
\newblock {\em Scientific reports}, 9(1):9237, 2019.

\bibitem{vlasic2008articulated}
D.~Vlasic, I.~Baran, W.~Matusik, and J.~Popovi{\'c}.
\newblock Articulated mesh animation from multi-view silhouettes.
\newblock In {\em ACM Transactions on Graphics (TOG)}, volume~27, page~97. ACM,
  2008.

\bibitem{xu2018powerful}
K.~Xu, W.~Hu, J.~Leskovec, and S.~Jegelka.
\newblock How powerful are graph neural networks?
\newblock {\em arXiv preprint arXiv:1810.00826}, 2018.

\end{thebibliography}
}

\end{document}